\documentclass[format=sigconf]{acmart}
\graphicspath{{fig/}}

\usepackage[utf8]{inputenc}
\usepackage{url}
\usepackage{amsmath,amsthm,amssymb,bm,mathtools}
\usepackage{enumitem}
\usepackage{booktabs}
\usepackage{siunitx}
\usepackage{caption}
\usepackage{subcaption}
\usepackage{tikz}
\usetikzlibrary{bayesnet}
\usepackage{balance} 

\usepackage{etoolbox}
\robustify\bfseries

\usepackage{algorithm}
\usepackage{algpseudocode}

\newcommand{\Emd}{\textemdash}
\newcommand{\GP}{\mathrm{GP}}
\newcommand{\KL}{\mathrm{KL}}
\newcommand{\Abs}[1]{\ensuremath{\lvert #1 \rvert}}
\newcommand{\Tr}{\top}  
\DeclareMathOperator*{\Argmax}{arg\,max}

\usepackage{xifthen}
\newcommand{\Exp}[2][]{\ensuremath{%
\ifthenelse{\isempty{#1}}{\mathbf{E}\left[ #2 \right]}{\mathbf{E}_{#1}\left[ #2 \right]}}}

\newlist{enuminline}{enumerate*}{1}
\setlist[enuminline,1]{label=\itshape\alph*\upshape)}

\copyrightyear{2019}
\acmYear{2019}
\setcopyright{acmlicensed}
\acmConference[KDD '19]{The 25th ACM SIGKDD Conference on Knowledge Discovery and Data Mining}{August 4--8, 2019}{Anchorage, AK, USA}
\acmBooktitle{The 25th ACM SIGKDD Conference on Knowledge Discovery and Data Mining (KDD '19), August 4--8, 2019, Anchorage, AK, USA}
\acmPrice{15.00}
\acmDOI{10.1145/3292500.3330831}
\acmISBN{978-1-4503-6201-6/19/08}

\begin{document}
\title{Pairwise Comparisons with Flexible Time-Dynamics}

\author{Lucas Maystre}
\authornote{This work was done while the author was at EPFL.}
\affiliation{%
  \institution{Spotify}
}
\email{lucasm@spotify.com}

\author{Victor Kristof}
\affiliation{%
  \institution{EPFL}
}
\email{victor.kristof@epfl.ch}

\author{Matthias Grossglauser}
\affiliation{%
  \institution{EPFL}
}
\email{matthias.grossglauser@epfl.ch}

\renewcommand{\shortauthors}{L. Maystre et al.}

\begin{abstract}
Inspired by applications in sports where the skill of players or teams competing against each other varies over time, we propose a probabilistic model of pairwise-comparison outcomes that can capture a wide range of time dynamics.
We achieve this by replacing the static parameters of a class of popular pairwise-comparison models by continuous-time Gaussian processes;
the covariance function of these processes enables expressive dynamics.
We develop an efficient inference algorithm that computes an approximate Bayesian posterior distribution.
Despite the flexbility of our model, our inference algorithm requires only a few linear-time iterations over the data and can take advantage of modern multiprocessor computer architectures.
We apply our model to several historical databases of sports outcomes and find that our approach
\begin{enuminline}
\item outperforms competing approaches in terms of predictive performance,
\item scales to millions of observations, and
\item generates compelling visualizations that help in understanding and interpreting the data.
\end{enuminline}

\end{abstract}

%
%


\maketitle

\section{Introduction}
\label{sec:intro}

In many competitive sports and games (such as tennis, basketball, chess and electronic sports), the most useful definition of a competitor's skill is the propensity of that competitor to win against an opponent.
It is often difficult to measure this skill \emph{explicitly}:
take basketball for example, a team's skill depends on the abilities of its players in terms of shooting accuracy, physical fitness, mental preparation, but also on the team's cohesion and coordination, on its strategy, on the enthusiasm of its fans, and a number of other intangible factors.
However, it is easy to observe this skill \emph{implicitly} through the outcomes of matches.

In this setting, probabilistic models of pairwise-comparison outcomes provide an elegant and practical approach to quantifying skill and to predicting future match outcomes given past data.
These models, pioneered by \citet{zermelo1928berechnung} in the context of chess (and by \citet{thurstone1927law} in the context of psychophysics), have been studied for almost a century.
They posit that each competitor $i$ (i.e., a team or player) is characterized by a latent score $s_i \in \mathbf{R}$ and that the outcome probabilities of a match between $i$ and $j$ are a function of the difference $s_i - s_j$ between their scores.
By estimating the scores $\{ s_i \}$ from data, we obtain an interpretable proxy for skill that is predictive of future match outcomes.
If a competitor's skill is expected to remain stable over time, these models are very effective.
But what if it varies over time?

A number of methods have been proposed to adapt comparison models to the case where scores change over time.
Perhaps the best known such method is the Elo rating system \citep{elo1978rating}, used by the World Chess Federation for their official rankings.
In this case, the time dynamics are captured essentially as a by-product of the learning rule (c.f. Section~\ref{sec:relwork}).
Other approaches attempt to model these dynamics explicitly \citep[e.g.,][]{fahrmeir1994dynamic, glickman1999parameter, dangauthier2007trueskill, coulom2008whole}.
These methods greatly improve upon the static case when considering historical data, but they all assume the simplest model of time dynamics (that is, Brownian motion).
Hence, they fail to capture more nuanced patterns such as variations at different timescales, linear trends, regression to the mean, discontinuities, and more.

In this work, we propose a new model of pairwise-comparison outcomes with expressive time-dynamics: it generalizes and extends previous approaches.
We achieve this by treating the score of an opponent $i$ as a time-varying Gaussian process $s_i(t)$ that can be endowed with flexible priors \citep{rasmussen2006gaussian}.
We also present an algorithm that, in spite of this increased flexibility, performs approximate Bayesian inference over the score processes in linear time in the number of observations so that our approach scales seamlessly to datasets with millions of observations.
This inference algorithm addresses several shortcomings of previous methods: it can be parallelized effortlessly and accommodates different variational objectives.
The highlights of our method are as follows.

\begin{description}
\item[Flexible Dynamics]
As scores are modeled by continuous-time Gaussian processes, complex (yet interpretable) dynamics can be expressed by composing covariance functions.

\item[Generality]
The score of an opponent for a given match is expressed as a (sparse) linear combination of features.
This enables, e.g., the representation of a home advantage or any other contextual effect.
Furthermore, the model encompasses a variety of observation likelihoods beyond win / lose, based, e.g., on the number of points a competitor scores.

\item[Bayesian Inference]
Our inference algorithm returns a posterior \emph{distribution} over score processes.
This leads to better predictive performance and enables a principled way to learn the dynamics (and any other model hyperparameters) by optimizing the log-marginal likelihood of the data.

\item[Ease of Intepretation]
By plotting the score processes $\{ s_i(t) \}$ over time, it is easy to visualize the probability of any comparison outcome under the model.
As the time dynamics are described through the composition of simple covariance functions, their interpretation is straightforward as well.
\end{description}

Concretely, our contributions are threefold.
First, we develop a probabilistic model of pairwise-comparison outcomes with flexible time-dynamics (Section~\ref{sec:model}).
The model covers a wide range of use cases, as it enables
\begin{enuminline}
\item opponents to be represented by a sparse linear combination of features, and
\item observations to follow various likelihood functions.
\end{enuminline}
In fact, it unifies and extends a large body of prior work.
Second, we derive an efficient algorithm for approximate Bayesian inference (Section~\ref{sec:inference}).
This algorithm adapts to two different variational objectives;
in conjunction with the ``reverse-KL'' objective, it provably converges to the optimal posterior approximation.
It can be parallelized easily, and the most computationally intensive step can be offloaded to optimized off-the-shelf numerical software.
Third, we apply our method on several sports datasets and show that it achieves state-of-the-art predictive performance (Section~\ref{sec:eval}).
Our results highlight that different sports are best modeled with different time-dynamics.
We also demonstrate how domain-specific and contextual information can improve performance even further;
in particular, we show that our model outperforms competing ones even when there are strong intransitivities in the data.

In addition to prediction tasks, our model can also be used to generate compelling visualizations of the temporal evolution of skills.
All in all, we believe that our method will be useful to data-mining practitioners interested in understanding comparison time-series and in building predictive systems for games and sports.

\paragraph{A Note on Extensions}
In this paper, we focus on \emph{pairwise} comparisons for conciseness.
However, the model and inference algorithm could be extended to multiway comparisons or partial rankings over small sets of opponents without any major conceptual change, similarly to \citet{herbrich2006trueskill}.
Furthermore, and even though we develop our model in the context of sports, it is relevant to all applications of ranking from comparisons, e.g., to those where comparison outcomes reflect human preferences or opinions \citep{thurstone1927law, mcfadden1973conditional, salganik2015wiki}.

\section{Model}
\label{sec:model}

In this section, we formally introduce our probabilistic model.
For clarity, we take a clean-slate approach and develop the model from scratch.
We discuss in more detail how it relates to prior work in Section~\ref{sec:relwork}.

The basic building blocks of our model are \emph{features}\footnote{%
In the simplest case, there is a one-to-one mapping between competitors (e.g., teams) and features, but decoupling them offers increased modeling power.}.
Let $M$ be the number of features; each feature $m \in [M]$ is characterized by a latent, continuous-time Gaussian process
\begin{align}
\label{eq:score}
s_m(t) \sim \GP[0, k_m(t, t')].
\end{align}
We call $s_m(t)$ the \emph{score process} of $m$, or simply its \emph{score}.
The \emph{covariance function} of the process, $k_m(t, t') \doteq \Exp{s_m(t) s_m(t')}$, is used to encode time dynamics.
A brief introduction to Gaussian processes as well as a discussion of useful covariance functions is given in Section~\ref{sec:covariances}.
The $M$ scores $s_1(t), \dots, s_M(t)$ are assumed to be (a priori) jointly independent, and we collect them into the \emph{score vector}
\begin{align*}
\bm{s}(t) = \begin{bmatrix}s_1(t) & \cdots & s_M(t) \end{bmatrix}^\Tr.
\end{align*}

For a given match, each opponent $i$ is described by a sparse linear combination of the features, with coefficients $\bm{x}_i \in \mathbf{R}^M$.
That is, the score of an opponent $i$ at time $t^*$ is given by
\begin{align}
\label{eq:compscore}
s_i = \bm{x}_i^\Tr \bm{s}(t^*).
\end{align}
In the case of a one-to-one mapping between competitors and features, $\bm{x}_i$ is simply the one-hot encoding of opponent $i$.
More complex setups are possible: For example, in the case of team sports and if the player lineup is available for each match, it could also be used to encode the players taking part in the match \citep{maystre2016player}.
Note that $\bm{x}_i$ can also depend contextually on the match.
For instance, it can be used to encode the fact that a team plays at home \citep{agresti2012categorical}.

Each observation consists of a tuple $(\bm{x}_i, \bm{x}_j, t^*, y)$, where $\bm{x}_i, \bm{x}_j$ are the opponents' feature vectors, $t^* \in \mathbf{R}$ is the time, and $y \in \mathcal{Y}$ is the match outcome.
We posit that this outcome is a random variable that depends on the opponents through their latent score difference:
\begin{align*}
y \mid \bm{x}_i, \bm{x}_j, t^* \sim p( y \mid s_i - s_j ),
\end{align*}
where $p$ is a known probability density (or mass) function and $s_i, s_j$ are given by~\eqref{eq:compscore}.
The idea of modeling outcome probabilities through score differences dates back to \citet{thurstone1927law} and \citet{zermelo1928berechnung}.
The likelihood $p$ is chosen such that positive values of $s_i - s_j$ lead to successful outcomes for opponent $i$ and vice-versa.

\begin{figure}[t]
  \subcaptionbox{
    Static model
  }[2cm]{
    \begin{tikzpicture}

\node[latent] (s)  {$s_m$};
\node[obs, below=of s] (y) {$y_n$};
\node[const, left=0.3cm of y, yshift=0.4cm] (x) {$\bm{x}_n$};

\edge {s} {y};
\edge[-] {x} {y};

\plate {scores} {(s)} {$M$}
\plate {observations} {(y)(x)} {$N$}
\end{tikzpicture}
  }
  \hfill
  \subcaptionbox{
    Our dynamic model \label{fig:model}
  }{
    \begin{tikzpicture}

\node[latent] (s1)  {$s_{m1}$};
\node[latent, right=1cm of s1] (s2)  {$s_{m2}$};
\node[latent, right=1.5cm of s2] (sn)  {$s_{mN}$};
\node[const, above=0.5cm of s1] (t1) {$t_1$};
\node[const, above=0.5cm of s2] (t2) {$t_2$};
\node[const, above=0.5cm of sn] (tn) {$t_N$};
\node[obs, below=of s1] (y1) {$y_1$};
\node[obs, below=of s2] (y2) {$y_2$};
\node[obs, below=of sn] (yn) {$y_N$};
\node[const, left=0.2cm of y1, yshift=0.4cm] (x1) {$\bm{x}_1$};
\node[const, left=0.3cm of y2, yshift=0.4cm] (x2) {$\bm{x}_2$};
\node[const, left=0.3cm of yn, yshift=0.4cm] (xn) {$\bm{x}_N$};

\edge[-] {t1} {s1};
\edge[-] {t2} {s2};
\edge[-] {tn} {sn};
\edge[-] {x1} {y1};
\edge[-] {x2} {y2};
\edge[-] {xn} {yn};
\edge {s1} {y1};
\edge {s2} {y2};
\edge {sn} {yn};

\path (t2) -- node[auto=false]{\ldots} (tn);
\path (y2) -- node[auto=false]{\ldots} (yn);
\draw[line width=2pt] (s1) -- (s2);
\draw[line width=2pt] (s2) -- node[fill=white] {\ldots} (sn);

\plate {scores} {(s1)(s2)(sn)} {$M$}
\end{tikzpicture}
  }
  \caption{
Graphical representation of a static model (left) and of the dynamic model presented in this paper (right).
The observed variables are shaded.
For conciseness, we let $\bm{x}_n \doteq \bm{x}_{n,i} - \bm{x}_{n,j}$.
Right: the latent score variables are mutually dependent across time, as indicated by the thick line.}
  \label{fig:pgms}
\end{figure}
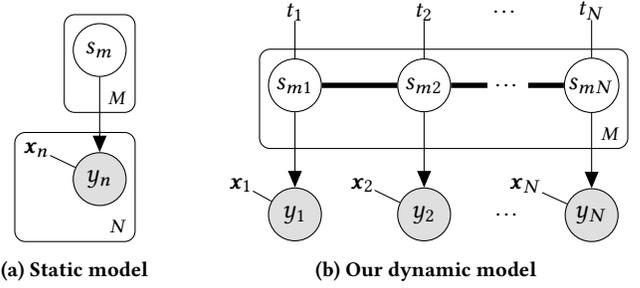

A graphical representation of the model is provided in Figure~\ref{fig:pgms}.
For perspective, we also include the representation of a static model, such as that of \citet{thurstone1927law}.
Our model can be interpreted as ``conditionally parametric'': conditioned on a particular time, it falls back to a (static) pairwise-comparison model parametrized by real-valued scores.

\paragraph{Observation Models}
Choosing an appropriate likelihood function $p(y \mid s_i - s_j)$ is an important modeling decision and depends on the information contained in the outcome $y$.
The most widely applicable likelihoods require only \emph{ordinal} observations, i.e., whether a match resulted in a win or a loss (or a tie, if applicable).
In some cases, we might additionally observe points (e.g., in association football, the number of goals scored by each team).
To make use of this extra information, we can model
\begin{enuminline}
\item the number of points of opponent $i$ with a Poisson distribution whose rate is a function of $s_i - s_j$, or
\item the points difference with a Gaussian distribution centered at $s_i - s_j$.
\end{enuminline}
A non-exhaustive list of likelihoods is given in Table~\ref{tab:likelihoods}.

\begin{table}[t]
  \caption{
Examples of observation likelihoods.
The score difference is denoted by $d \doteq s_i - s_j$ and the Gaussian cumulative density function is denoted by $\Phi$.
}
  \label{tab:likelihoods}
  \centering
  \begin{tabular}{l lll}
    \toprule
    Name           & $\mathcal{Y}$        & $p(y \mid d)$               & References \\
    \midrule
    Probit         & $\{\pm 1 \}$         & $\Phi(yd)$                  & \citep{thurstone1927law, herbrich2006trueskill}   \\
    Logit          & $\{\pm 1 \}$         & $[1 + \exp(-yd)]^{-1}$      & \citep{zermelo1928berechnung, bradley1952rank}    \\
    Ordinal probit & $\{\pm 1, 0 \}$      & $\Phi(yd - \alpha), \ldots$ & \citep{glenn1960ties}      \\
    Poisson-exp    & $\mathbf{N}_{\ge 0}$ & $\exp(yd - e^d) / y!$       & \citep{maher1982modelling} \\
    Gaussian       & $\mathbf{R}$         & $\propto \exp[(y - d)^2 / (2\sigma^2)]$ & \citep{guo2012score} \\
    \bottomrule
  \end{tabular}
\end{table}

\subsection{Covariance Functions}
\label{sec:covariances}

A Gaussian process $s(t) \sim \GP[0, k(t, t')]$ can be thought of as an infinite collection of random variables indexed by time, such that the joint distribution of any finite vector of $N$ samples $\bm{s} = [s(t_1) \cdots s(t_N)]$ is given by $\bm{s} \sim \mathcal{N}(\bm{0}, \bm{K})$, where $\bm{K} = [k(t_i, t_j)]$.
That is, $\bm{s}$ is jointly Gaussian with mean $\bm{0}$ and covariance matrix $\bm{K}$.
We refer the reader to \citet{rasmussen2006gaussian} for an excellent introduction to Gaussian processes.

\begin{figure*}[t]
  \includegraphics{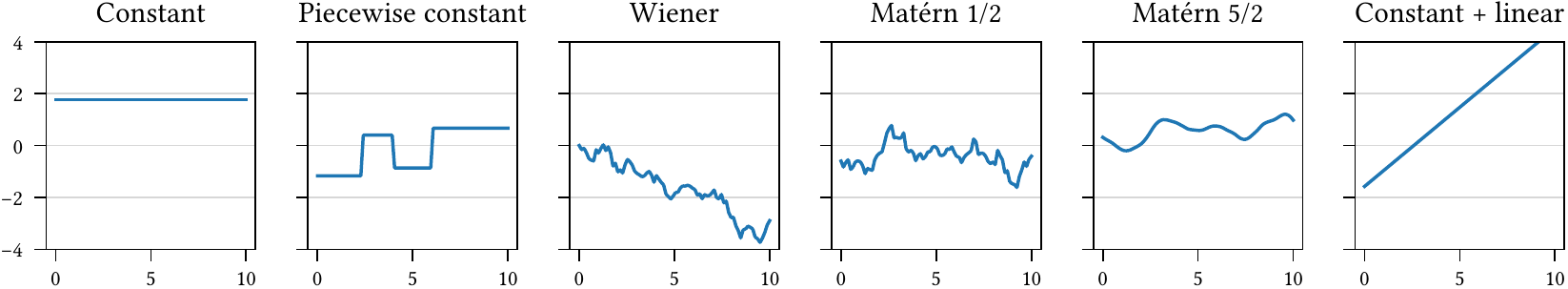}
  \caption{Random realizations of a zero-mean Gaussian process with six different covariance functions.}
  \label{fig:covariances}
\end{figure*}

Hence, by specifying the covariance function appropriately, we can express prior expectations about the time dynamics of a feature's score, such as smooth or non-smooth variations at different timescales, regression to the mean, discontinuities, linear trends and more.
Here, we describe a few functions that we find useful in the context of modeling temporal variations.
Figure~\ref{fig:covariances} illustrates these functions through random realizations of the corresponding Gaussian processes.

\begin{description}
\item[Constant] This covariance captures processes that remain constant over time.
It is useful in composite covariances to model a constant offset (i.e., a mean score value).

\item[Piecewise Constant]
Given a partition of $\mathbf{R}$ into disjoint intervals, this covariance is constant inside a partition and zero between partitions.
It can, for instance, capture discontinuities across seasons in professional sports leagues.

\item[Wiener] This covariance reflects Brownian motion dynamics (c.f. Section~\ref{sec:relwork}).
It is non-stationary: the corresponding process drifts away from $0$ as $t$ grows.

\item[Matérn] This family of stationary covariance functions can represent smooth and non-smooth variations at various timescales.
It is parametrized by a variance, a characteristic timescale and a smoothness parameter $\nu$.
When $\nu = 1/2$, it corresponds to a mean-reverting version of Brownian motion.

\item[Linear] This covariance captures linear dynamics.
\end{description}

Finally, note that composite functions can be created by adding or multiplying covariance functions together.
For example, let $k_a$ and $k_b$ be constant and Matérn covariance functions, respectively.
Then, the composite covariance $k(t, t') \doteq k_a(t, t') + k_b(t, t')$ captures dynamics that fluctuate around a (non-zero) mean value.
\citet[Sec. 2.3]{duvenaud2014automatic} provides a good introduction to building expressive covariance functions by composing simple ones.

\section{Inference Algorithm}
\label{sec:inference}

In this section, we derive an efficient inference algorithm for our model.
For brevity, we focus on explaining the main ideas behind the algorithm.
A reference software implementation, available online at \url{https://github.com/lucasmaystre/kickscore}, complements the description provided here.

We begin by introducing some notation.
Let $\mathcal{D} = \{ (\bm{x}_n, t_n, y_n) : n \in [N] \}$ be a dataset of $N$ independent observations, where for conciseness we fold the two opponents $\bm{x}_{n,i}$ and $\bm{x}_{n,j}$ into $\bm{x}_n \doteq \bm{x}_{n,i} - \bm{x}_{n,j}$, for each observation\footnote{%
This enables us to write the score difference more compactly.
Given an observation at time $t^*$ and letting $\bm{x} \doteq \bm{x}_i - \bm{x}_j$, we have $s_i - s_j = \bm{x}_i^\Tr \bm{s}(t^*) - \bm{x}_j^\Tr \bm{s}(t^*) = \bm{x}^\Tr \bm{s}(t^*)$.
}.
Let $\mathcal{D}_m \subseteq [N]$ be the subset of observations involving feature $m$, i.e., those observations for which $x_{nm} \ne 0$, and let $N_m = \lvert \mathcal{D}_m \rvert$.
Finally, denote by $\bm{s}_m \in \mathbf{R}^{N_m}$ the samples of the latent score process at times corresponding to the observations in $\mathcal{D}_m$.
The joint prior distribution of these samples is $p(\bm{s}_m) = \mathcal{N}(\bm{0}, \bm{K}_m)$, where $\bm{K}_m$ is formed by evaluating the covariance function $k_m(t, t')$ at the relevant times.

We take a Bayesian approach and seek to compute the posterior distribution
\begin{align}
\label{eq:truepost}
p(\bm{s}_1, \ldots, \bm{s}_M \mid \mathcal{D}) \propto \prod_{m = 1}^M p(\bm{s}_m) \prod_{n = 1}^N p[y_n \mid \bm{x}_n^\Tr \bm{s}(t_n)].
\end{align}
As the scores are coupled through the observations, the posterior no longer factorizes over $\{ \bm{s}_m \}$.
Furthermore, computing the posterior is intractable if the likelihood is non-Gaussian.

To overcome these challenges, we consider a mean-field variational approximation \citep{wainwright2008graphical}.
In particular, we assume that the posterior can be well-approximated by a multivariate Gaussian distribution that factorizes over the features:
\begin{align}
\label{eq:approxpost}
p(\bm{s}_1, \ldots, \bm{s}_M \mid \mathcal{D})
    \approx q(\bm{s}_1, \ldots, \bm{s}_M)
    \doteq \prod_{m = 1}^M \mathcal{N}(\bm{s}_m \mid \bm{\mu}_m, \bm{\Sigma}_m).
\end{align}
Computing this approximate posterior amounts to finding the variational parameters $\{\bm{\mu}_m, \bm{\Sigma}_m \}$ that best approximate the true posterior.
More formally, the inference problem reduces to the optimization problem
\begin{align}
\label{eq:varopt}
\min_{\{\bm{\mu}_m, \bm{\Sigma}_m \}} \mathrm{div} \left[ p(\bm{s}_1, \ldots, \bm{s}_M \mid \mathcal{D}) \;\Vert\; q(\bm{s}_1, \ldots, \bm{s}_M) \right],
\end{align}
for some divergence measure $\mathrm{div}(p \Vert q) \ge 0$.
We will consider two different such measures in Section~\ref{sec:inf-pseudo-obs}.

A different viewpoint on the approximate posterior is as follows.
For both of the variational objectives that we consider, it is possible to rewrite the optimal distribution $q(\bm{s}_m)$ as
\begin{align*}
q(\bm{s}_m) \propto p(\bm{s}_m) \prod_{n \in \mathcal{D}_m} \mathcal{N}[s_{mn} \mid \tilde{\mu}_{mn}, \tilde{\sigma}^2_{mn}].
\end{align*}
Letting $\mathcal{X}_n \subseteq [M]$ be the subset of features such that $x_{nm} \ne 0$, we can now reinterpret the variational approximation as transforming every observation $(\bm{x}_n, t_n, y_n)$ into several independent \emph{pseudo-observations} with Gaussian likelihood, one for each feature $m \in \mathcal{X}_n$.
Instead of optimizing directly $\{ \bm{\mu}_m, \bm{\Sigma}_m \}$ in~\eqref{eq:varopt}, we can alternatively choose to optimize the parameters $\{ \tilde{\mu}_{mn}, \tilde{\sigma}^2_{mn} \}$.
For any feature $m$, given the pseudo-observations' parameters $\tilde{\bm{\mu}}_m$ and $\tilde{\bm{\sigma}}_m^2$, computing $q(\bm{s}_m)$ becomes tractable (c.f. Section~\ref{sec:inf-posterior}).

An outline of our iterative inference procedure is given in Algorithm~\ref{alg:inference}.
Every iteration consists of two steps:
\begin{enumerate}
\item updating the pseudo-observations' parameters given the true observations and the current approximate posterior (lines \ref{li:begin-param}--\ref{li:end-param}), and
\item recomputing the approximate posterior given the current pseudo-observation (lines \ref{li:begin-post} and \ref{li:end-post}).
\end{enumerate}
Convergence is declared when the difference between two successive iterates of $\{ \tilde{\mu}_{mn} \}$ and $\{ \tilde{\sigma}_{mn}^2 \}$ falls below a threshold.
Note that, as a by-product of the computations performed by the algorithm, we can also estimate the log-marginal likelihood of the data, $\log p(\mathcal{D})$.

\algtext*{EndFor}
\begin{algorithm}[t]
  \caption{Model inference.}
  \label{alg:inference}
  \begin{algorithmic}[1]
    \Require $\mathcal{D} = \{ (\bm{x}_n, t_n, y_n) : n \in [N] \}$
    \State $\tilde{\bm{\mu}}_m, \tilde{\bm{\sigma}}^2_m \gets \bm{0}, \bm{\infty} \quad \forall m$
    \State $q(\bm{s}_m) \gets p(\bm{s}_m) \quad \forall m$
    \Repeat
      \For{$n = 1, \ldots, N$}  \label{li:begin-param}
        \State $\bm{\delta} \gets \textsc{Derivatives}(\bm{x}_n, y_n)$ \label{li:derivatives}
        \For{$m \in \mathcal{X}_n$}
          \State $\tilde{\mu}_{mn}, \tilde{\sigma}^2_{mn} \gets \textsc{UpdateParams}(x_{nm}, \bm{\delta})$ \label{li:updateparams}
        \EndFor
      \EndFor \label{li:end-param}
      \For{$m = 1, \ldots, M$} \label{li:begin-post}
        \State $q(\bm{s}_m) \gets \textsc{UpdatePosterior}(\tilde{\bm{\mu}}_m, \tilde{\bm{\sigma}}^2_m)$ \label{li:updateposterior}
      \EndFor \label{li:end-post}
    \Until convergence
  \end{algorithmic}
\end{algorithm}

\paragraph{Running Time}
In Appendix~\ref{app:inference}, we show that \textsc{Derivatives} and \textsc{UpdateParams} run in constant time.
In Section~\ref{sec:inf-posterior}, we show that \textsc{UpdatePosterior} runs in time $O(N_m)$.
Therefore, if we assume that the vectors $\{ \bm{x}_n \}$ are sparse, the total running time per iteration of Algorithm~\ref{alg:inference} is $O(N)$.
Furthermore, each of the two outer \emph{for} loops (lines \ref{li:begin-param} and \ref{li:begin-post}) can be parallelized easily, leading in most cases to a linear acceleration with the number of available processors.

\subsection{Updating the Pseudo-Observations}
\label{sec:inf-pseudo-obs}

The exact computations performed during the first step of the inference algorithm---updating the pseudo-observations---depend on the specific variational method used.
We consider two: expectation propagation \citep{minka2001family}, and reverse-KL variational inference~\citep{blei2017variational}.
The ability of Algorithm~\ref{alg:inference} to seamlessly adapt to either of the two methods is valuable, as it enables practitioners to use the most advantageous method for a given likelihood function.
Detailed formulae for \textsc{Derivatives} and \textsc{UpdateParams} can be found in Appendix~\ref{app:inference}.

\subsubsection{Expectation Propagation}

We begin by defining two distributions.
The \emph{cavity} distribution $q_{-n}$ is the approximate posterior without the pseudo-observations associated with the $n$th datum, that is,
\begin{align*}
q_{-n}(\bm{s}_1, \ldots, \bm{s}_M) \propto \frac{q(\bm{s}_1, \ldots, \bm{s}_M)}{
        \prod_{m \in \mathcal{X}_n} \mathcal{N}[s_{mn} \mid \tilde{\mu}_{mn}, \tilde{\sigma}^2_{mn}]}.
\end{align*}
The \emph{hybrid} distribution $\hat{q}_n$ is given by the cavity distribution multiplied by the $n$th likelihood factor, i.e.,
\begin{align*}
\hat{q}_n(\bm{s}_1, \ldots, \bm{s}_M) \propto q_{-n}(\bm{s}_1, \ldots, \bm{s}_M) p[y_n \mid \bm{x}_n^\Tr \bm{s}(t_n)].
\end{align*}
Informally, the hybrid distribution $\hat{q}_n$ is ``closer'' to the true distribution than $q$.

Expectation propagation (EP) works as follows. At each iteration and for each $n$, we update the parameters $\{ \tilde{\mu}_{mn}, \tilde{\sigma}_{mn} : m \in \mathcal{X}_n \}$ such that $\KL( \hat{q}_n \Vert q )$ is minimized.
To this end, the function \textsc{Derivatives} (on line~\ref{li:derivatives} of Algorithm~\ref{alg:inference}) computes the first and second derivatives of the log-partition function
\begin{align}
\label{eq:logpart}
\log \mathbf{E}_{q_{-n}} \left\{ p[y_n \mid \bm{x}_n^\Tr \bm{s}(t_n)] \right\}
\end{align}
with respect to ${\mu}_{-n} \doteq \mathbf{E}_{q_{-n}}[\bm{x}_n^\Tr \bm{s}(t_n)]$.
These computations can be done in closed form for the widely-used probit likelihood, and they involve one-dimensional numerical integration for most other likelihoods.
EP has been reported to result in more accurate posterior approximations on certain classification tasks \citep{nickisch2008approximations}.

\subsubsection{Reverse KL Divergence}

This method (often referred to simply as \emph{variational inference} in the literature) seeks to minimize $\KL(q \Vert p)$, i.e., the KL divergence from the approximate posterior $q$ to the true posterior $p$.

To optimize this objective, we adopt the approach of \citet{khan2017conjugate}.
In this case, the function \textsc{Derivatives} computes the first and second derivatives of the expected log-likelihood
\begin{align}
\label{eq:exp-ll}
\mathbf{E}_q \left\{ \log p[y_n \mid \bm{x}_n^\Tr \bm{s}(t_n)] \right\}
\end{align}
with respect to $\mu \doteq \mathbf{E}_q[\bm{x}_n^\Tr \bm{s}(t_n)]$.
These computations involve numerically solving two one-dimensional integrals.

In comparison to EP, this method has two advantages.
The first is theoretical:
If the likelihood $p(y \mid d)$ is log-concave in $d$, then the variational objective has a unique global minimum, and we can guarantee that Algorithm~\ref{alg:inference} converges to this minimum \citep{khan2017conjugate}.
The second is numerical:
Excepted for the probit likelihood, computing~\eqref{eq:exp-ll} is numerically more stable than computing~\eqref{eq:logpart}.

\subsection{Updating the Approximate Posterior}
\label{sec:inf-posterior}

The second step of Algorithm~\ref{alg:inference} (lines~\ref{li:begin-post} and~\ref{li:end-post}) solves the following problem, for every feature $m$.
Given Gaussian pseudo-observations $\{ \tilde{\mu}_{mn}, \tilde{\sigma}_{mn} : n \in \mathcal{D}_m \}$ and a Gaussian prior $p(\bm{s}_m) = \mathcal{N}(\bm{0}, \bm{K}_m)$, compute the posterior
\begin{align*}
q(\bm{s}_m) \propto p(\bm{s}_m) \prod_{n \in \mathcal{D}_m} \mathcal{N}[s_{mn} \mid \tilde{\mu}_{mn}, \tilde{\sigma}^2_{mn}].
\end{align*}
This computation can be done independently and in parallel for each feature $m \in [M]$.

A naive approach is to use the self-conjugacy properties of the Gaussian distribution directly.
Collecting the parameters of the pseudo-observations into a vector $\tilde{\bm{\mu}}_m$ and a diagonal matrix $\tilde{\bm{\Sigma}}_m$, the parameters of the posterior $q(\bm{s}_m)$ are given by
\begin{align}
\label{eq:batch}
\bm{\Sigma}_m = (\bm{K}_m^{-1} + \tilde{\bm{\Sigma}}_m^{-1})^{-1}, \qquad
\bm{\mu}_m    = \bm{\Sigma}_m \tilde{\bm{\Sigma}}_m^{-1} \tilde{\bm{\mu}}_m.
\end{align}
Unfortunately, this computation runs in time $O(N_m^3)$, a cost that becomes prohibitive if some features appear in many observations.

Instead, we use an alternative approach that exploits a link between temporal Gaussian processes and state-space models \citep{hartikainen2010kalman, reece2010introduction}.
Without loss of generality, we now assume that the $N$ observations are ordered chronologically, and, for conciseness, we drop the feature's index and consider a single process $s(t)$.
The key idea is to augment $s(t)$ into a $K$-dimensional vector-valued Gauss-Markov process $\bar{\bm{s}}(t)$, such that
\begin{align*}
\bar{\bm{s}}(t_{n+1}) = \bm{A}_n \bar{\bm{s}}(t_n) + \bm{\varepsilon}_n,
    \qquad \bm{\varepsilon}_n \sim \mathcal{N}(\bm{0}, \bm{Q}_n)
\end{align*}
where $K \in \mathbf{N}_{>0}$ and $\bm{A}_n, \bm{Q}_n \in \mathbf{R}^{K \times K}$ depend on the time interval $\Abs{t_{n+1} - t_n}$ and on the covariance function $k(t, t')$ of the original process $s(t)$.
The original (scalar-valued) and the augmented (vector-valued) processes are related through the equation
\begin{align*}
s(t) = \bm{h}^\Tr \bar{\bm{s}}(t),
\end{align*}
where $\bm{h} \in \mathbf{R}^K$ is called the \emph{measurement vector}.

\begin{figure}[t]
  \centering
  \begin{tikzpicture}

\node[latent] (s1)  {$\bar{s}_{m1}^k$};
\node[latent, right=1cm of s1] (s2)  {$\bar{s}_{m2}^k$};
\node[latent, right=1.5cm of s2] (sn)  {$\bar{s}_{mN}^k$};
\node[const, above=0.5cm of s1] (t1) {$t_1$};
\node[const, above=0.5cm of s2] (t2) {$t_2$};
\node[const, above=0.5cm of sn] (tn) {$t_N$};
\node[obs, below=1.5cm of s1] (y1) {$y_1$};
\node[obs, below=1.5cm of s2] (y2) {$y_2$};
\node[obs, below=1.5cm of sn] (yn) {$y_N$};
\node[const, left=0.2cm of y1, yshift=0.4cm] (x1) {$\bm{x}_1$};
\node[const, left=0.3cm of y2, yshift=0.4cm] (x2) {$\bm{x}_2$};
\node[const, left=0.3cm of yn, yshift=0.4cm] (xn) {$\bm{x}_N$};

\edge[-] {t1} {s1};
\edge[-] {t2} {s2};
\edge[-] {tn} {sn};
\edge[-] {x1} {y1};
\edge[-] {x2} {y2};
\edge[-] {xn} {yn};
\edge {s1} {y1};
\edge {s2} {y2};
\edge {sn} {yn};
\edge {s1} {s2};

\path (t2) -- node[auto=false]{\ldots} (tn);
\path (y2) -- node[auto=false]{\ldots} (yn);
\path (s2) edge[->, >={triangle 45}] node[fill=white] {\ldots} (sn);

\plate {derivs1} {(s1)} {$K$}
\plate {derivs2} {(s2)} {$K$}
\plate {derivsn} {(sn)} {$K$}
\plate {scores} {(derivs1)(derivs2)(derivsn)} {$M$}
\end{tikzpicture}
  \caption{State-space reformulation of our model.
  With respect to the representation in Figure~\ref{fig:model}, the number of latent variables has increased, but they now form a Markov chain.}
  \label{fig:ssm}
\end{figure}
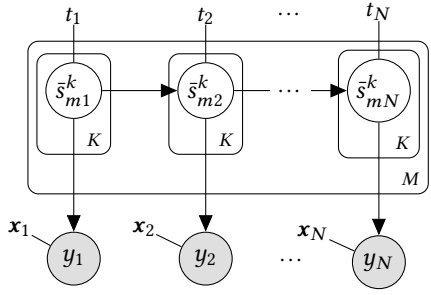

Figure~\ref{fig:ssm} illustrates our model from a state-space viewpoint.
It is important to note that the mutual time dependencies of Figure~\ref{fig:model} have been replaced by Markovian dependencies.
In this state-space formulation, posterior inference can be done in time $O(K^3 N)$ by using the Rauch--Tung--Striebel smoother \citep{sarkka2013bayesian}.

\paragraph{From Covariance Functions to State-Space Models}
A method for converting a process $s(t) \sim \GP[\bm{0}, k(t, t')]$ into an equivalent Gauss-Markov process $\bar{\bm{s}}(t)$ by explicit construction of $\bm{h}$, $\{\bm{A}_n\}$ and $\{\bm{Q}_n\}$ is given in \citet{solin2016stochastic}.
All the covariance functions described in Section~\ref{sec:covariances} lead to exact state-space reformulations of order $K \leq 3$.
The composition of covariance functions through addition or multiplication can also be treated exactly and automatically.
Some other covariance functions, such as the squared-exponential function or periodic functions \citep{rasmussen2006gaussian}, cannot be transformed exactly but can be approximated effectively and to arbitrary accuracy \citep{hartikainen2010kalman, solin2014explicit}.

Finally, we stress that the state-space viewpoint is useful because it leads to a faster inference procedure; but defining the time dynamics of the score processes in terms of covariance functions is much more intuitive.

\subsection{Predicting at a New Time}

Given the approximate posterior $q(\bm{s}_1, \ldots, \bm{s}_M)$, the probability of observing outcome $y$ at a new time $t^*$ given the feature vector $\bm{x}$ is given by
\begin{align*}
p(y \mid \bm{x}, t^*) = \int_\mathbf{R} p(y \mid z) p(z) dz,
\end{align*}
where $z = \bm{x}^\Tr \bm{s}(t^*)$ and the distribution of $s_m(t^*)$ is derived from the posterior $q(\bm{s}_m)$.
By using the state-space formulation of the model, the prediction can be done in constant time \citep{saatci2012scalable}.

\section{Experimental Evaluation}
\label{sec:eval}

\begin{figure*}[t]
  \includegraphics{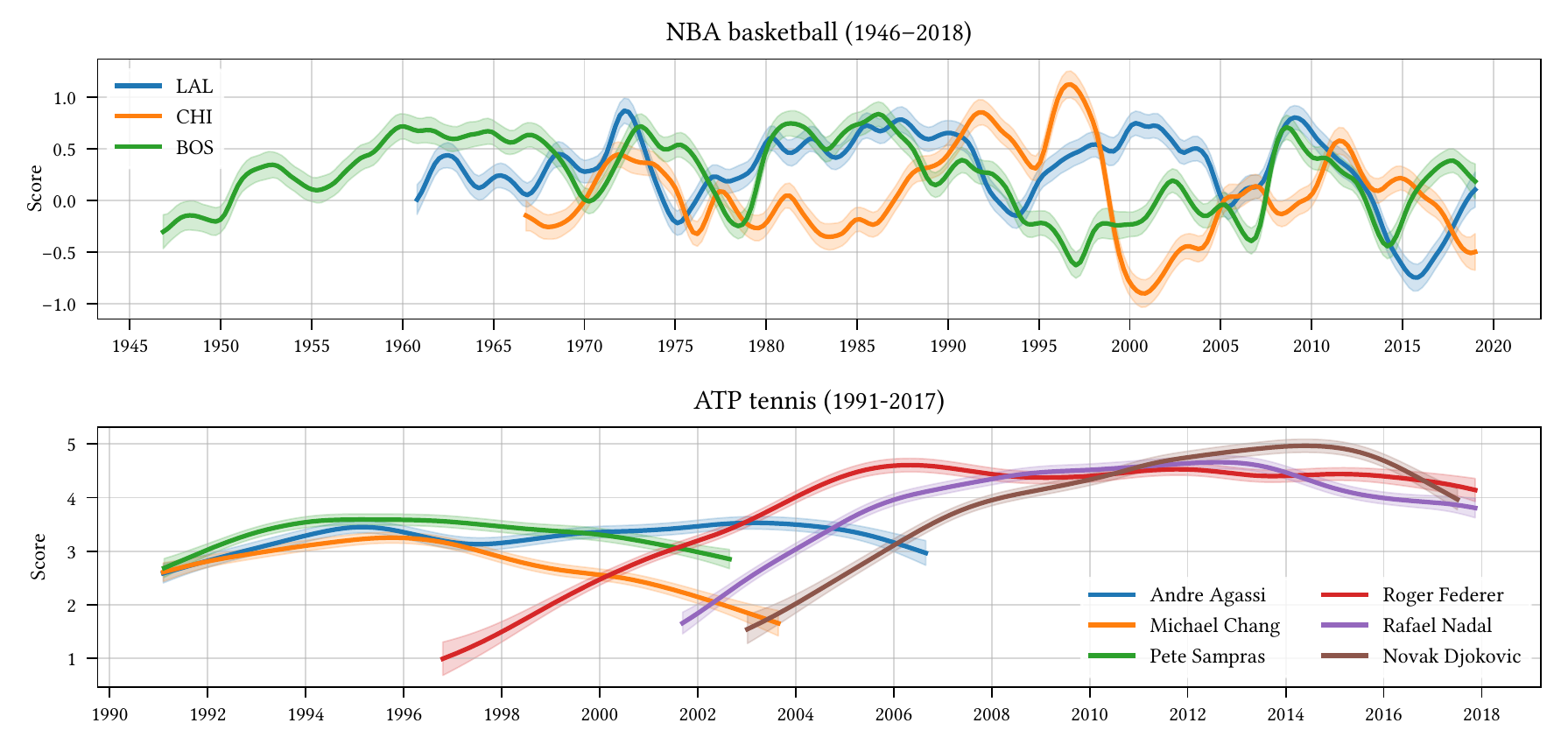}
  \caption{
Temporal evolution of the score processes ($\mu \pm \sigma$) corresponding to selected basketball teams (top) and tennis players (bottom).
The basketball teams are the Los Angeles Lakers (LAL), the Chicago Bulls (CHI) and the Boston Celtics (BOS).}
  \label{fig:scores}
\end{figure*}

In this section, we evaluate our model and inference algorithm on real data.
Our experiments cover three aspects.
First, in Section~\ref{sec:evaldyn}, we compare the predictive performance of our model against competing approaches, focusing on the impact of flexible time-dynamics.
Second, in Section~\ref{sec:evalgen}, we show that by carefully choosing features and observation likelihoods, predictive performance can be improved significantly.
Finally, in Section~\ref{sec:evalinf}, we study various facets of our inference algorithm.
We measure the impact of the mean-field assumption and of the choice of variational objective, and we demonstrate the scalability of the algorithm.

\paragraph{Datasets}
We consider six datasets of pairwise-comparison outcomes of various sports and games.
Four of them contain timestamped outcomes; they relate to tennis, basketball, association football and chess.
Due to the large size of the chess dataset\footnote{%
This dataset consists of all the match outcomes contained in \emph{ChessBase Big Database 2018}, available at \url{https://shop.chessbase.com/en/products/big_database_2018}.}, we also consider a subset of the data spanning 30 years.
The two remaining datasets contain match outcomes of the StarCraft computer game and do not have timestamps.
Table~\ref{tab:datasets} provides summary statistics for all the datasets.
Except for chess, all data are publicly available online\footnote{%
Tennis: \url{https://github.com/JeffSackmann/tennis_atp},
basketball: \url{https://projects.fivethirtyeight.com/nba-model/nba_elo.csv},
football: \url{https://int.soccerway.com/},
StarCraft: \url{https://github.com/csinpi/blade_chest}.
}.

\begin{table}[t]
  \caption{
Summary statistics of the sports datasets.}
  \label{tab:datasets}
  \centering
  \sisetup{table-text-alignment=right}
\begin{tabular}{l lS[table-format=6.0]S[table-format=7.0]l}
  \toprule
  Name            & Ties &    $M$ &     $N$ &  Time span \\
  \midrule
  ATP tennis      & No   &  20046 &  618934 & 1991--2017 \\
  NBA basketball  & No   &    102 &   67642 & 1946--2018 \\
  World football  & Yes  &    235 &   19158 & 1908--2018 \\
  ChessBase small & Yes  &  19788 &  306764 & 1950--1980 \\
  ChessBase full  & Yes  & 343668 & 7169202 & 1475--2017 \\
  \addlinespace
  StarCraft WoL   & No   &   4381 &   61657 & ---        \\
  StarCraft HotS  & No   &   2287 &   28582 & ---        \\
  \bottomrule
\end{tabular}

\end{table}

\paragraph{Performance Metrics}
Let $(\bm{x}, t^*, y)$ be an observation.
We measure performance by using
the logarithmic loss: $-\log p(y \mid \bm{x}, t^*)$ and
the accuracy: $\textbf{1}\{y = \Argmax_{y'} p(y' \mid \bm{x}, t^*)\}$.
We report their average values on the test set.

\paragraph{Methodology}
Unless specified otherwise, we partition every dataset into a training set containing the first 70\% of the observations and a test set containing the remaining 30\%, in chronological order.
The various hyperparameters (such as covariance functions and their parameters, learning rates, etc.) are selected based on the training data only, by maximizing the log-marginal likelihood of Bayesian models and by minimizing the average leave-one-out log loss otherwise.
The final hyperparameter configuration of all models can be found in Appendix~\ref{app:eval}.
In order to predict the outcome of an observation at time $t^*$, we use \emph{all} the data (in both training and test sets) up to the day preceding $t^*$.
This closely mimics the setting where a predictor must guess the outcome of an event in the near future based on all past data.
Unless specified otherwise, we use Algorithm~\ref{alg:inference} with the EP variational objective, and we declare convergence when the improvement in log-marginal likelihood falls below $10^{-3}$.
Typically, the algorithm converges in less than a hundred iterations.

\subsection{Flexible Time-Dynamics}
\label{sec:evaldyn}

In this experiment, we compare the predictive performance of our model against competing approaches on four timestamped datasets.
In order to better isolate and understand the impact of accurately modeling \emph{time dynamics} on predictive performance, we keep the remaining modeling choices simple: we treat all outcomes as ordinal-valued (i.e., \emph{win}, \emph{loss} and possibly \emph{tie}) with a probit likelihood and use a one-to-one mapping between competitors and features.
In Table~\ref{tab:predperf}, we report results for the following models:
\begin{itemize}
\item \emph{Random}. This baseline assigns equal probability to every outcome.

\item \emph{Constant}. The model of Section~\ref{sec:model} with a constant covariance function.
This model assumes that the scores do not vary over time.

\item \emph{Elo}. The system used by the World Chess Federation \citep{elo1978rating}.
Time dynamics are a by-product of the update rule (c.f. Section~\ref{sec:relwork}).

\item \emph{TrueSkill}. The Bayesian model of \citet{herbrich2006trueskill}.
Time dynamics are assumed to follow Brownian motion (akin to our Wiener kernel) and inference is done in a single pass over the data.

\item \emph{Ours}. The model of Section~\ref{sec:model}.
We try multiple covariance functions and report the one that maximizes the log-marginal likelihood.
\end{itemize}

\begin{table*}[t]
  \caption{
Predictive performance of our model and of competing approaches on four datasets, in terms of average log loss and average accuracy.
The best result is indicated in bold.}
  \label{tab:predperf}
  \centering
  \sisetup{table-format=1.3, detect-all}
\begin{tabular}{l SS SS SS SS SSl}
  \toprule
                  & \multicolumn{2}{c}{Random}    & \multicolumn{2}{c}{Constant}  & \multicolumn{2}{c}{Elo}
                      & \multicolumn{2}{c}{TrueSkill} & \multicolumn{3}{c}{Ours} \\
                    \cmidrule(r){2-3}               \cmidrule(r){4-5}               \cmidrule(r){6-7}
                        \cmidrule(r){8-9}               \cmidrule{10-12}
  Dataset         &          Loss &   \text{Acc.} &          Loss &   \text{Acc.} &          Loss &   \text{Acc.}
                      &          Loss &   \text{Acc.} &          Loss &   \text{Acc.} & Covariance \\
  \midrule
  ATP tennis      &         0.693 &        0.500 &         0.581 &         0.689 &         0.563 &        0.705
                      &         0.563 &         0.705 & \bfseries 0.552 & \bfseries 0.714 &  Affine + Wiener \\
  NBA basketball  &         0.693 &         0.500 &         0.692 &         0.536 &         0.634 &         0.644
                      &         0.634 &         0.644 & \bfseries 0.630 & \bfseries 0.645 &  Constant + Matérn 1/2 \\
  World football  &         1.099 &         0.333 &         0.929 & \bfseries 0.558 &         0.950 &         0.551
                      &         0.937 &         0.554 & \bfseries 0.926 & \bfseries 0.558 &  Constant + Matérn 1/2 \\
  ChessBase small &         1.099 &         0.333 &         1.030 & \bfseries 0.478 &         1.035 &         0.447
                      &         1.030 &         0.467 & \bfseries 1.026 &           0.474 &  Constant + Wiener \\
  \bottomrule
\end{tabular}

\end{table*}

Our model matches or outperforms other approaches in almost all cases, both in terms of log loss and in terms of accuracy.
Interestingly, different datasets are best modeled by using different covariance functions, perhaps capturing underlying skill dynamics specific to each sport.

\paragraph{Visualizing and Interpreting Scores}
Figure~\ref{fig:scores} displays the temporal evolution of the score of selected basketball teams and tennis players.
In the basketball case, we can recognize the dominance of the Boston Celtics in the early 1960's and the Chicago Bulls' strong 1995-96 season.
In the tennis case, we can see the progression of a new generation of tennis champions at the turn of the 21\textsuperscript{st} century.
Plotting scores over time provides an effective way to compactly represent the history of a given sport.
Analyzing the optimal hyperparameters (c.f. Table~\ref{tab:hpvals} in Appendix~\ref{app:eval}) is also insightful: the characteristic timescale of the dynamic covariance component is \num{1.75} and \num{7.47} years for basketball and tennis, respectively.
The score of basketball teams appears to be much more volatile.

\subsection{Generality of the Model}
\label{sec:evalgen}

In this section, we demonstrate how we can take advantage of additional modeling options to further improve predictive performance.
In particular, we show that
choosing an appropriate likelihood and
parametrizing opponents with match-dependent combinations of features
can bring substantial gains.

\subsubsection{Observation Models}
Basketball and football match outcomes actually consist of \emph{points} (respectively, goals) scored by each team during the match.
We can make use of this additional information to improve predictions \citep{maher1982modelling}.
For each of the basketball and football datasets, we compare the best model obtained in Section~\ref{sec:evaldyn} to alternative models.
These alternative models keep the same time dynamics but use either
\begin{enumerate}
\item a logit likelihood on the ordinal outcome,
\item a Gaussian likelihood on the points difference, or
\item a Poisson-exp likelihood on the points scored by each team.
\end{enumerate}
The results are presented in Table~\ref{tab:lklperf}.
The logit likelihood performs similarly to the probit one \citep{stern1992all}, but likelihoods that take points into account can indeed lead to better predictions.

\begin{table}[ht]
  \caption{
Average predictive log loss of models with different observation likelihoods.
The best result is indicated in bold.}
  \label{tab:lklperf}
  \centering
  \sisetup{table-format=1.3, detect-all}
\begin{tabular}{l SSSS}
  \toprule
  Dataset         & \text{Probit} &  \text{Logit} & \text{Gaussian}  &  \text{Poisson} \\
  \midrule
  NBA basketball  &         0.630 &         0.630 &  \bfseries 0.627 &           0.630 \\
  World football  &         0.926 &         0.926 &           0.927  & \bfseries 0.922 \\
  \bottomrule
\end{tabular}

\end{table}

\subsubsection{Match-Dependent Parametrization}
For a given match, we can represent opponents by using (non-trivial) linear combinations of features.
This enables, e.g., to represent context-specific information that might influence the outcome probabilities.
In the case of football, for example, it is well-known that a team playing at home has an advantage.
Similarly, in the case of chess, playing White results in a slight advantage.
Table~\ref{tab:homeadv} displays the predictive performance achieved by our model when the score of the home team (respectively, that of the opponent playing White) is modeled by a linear combination of two features: the identity of the team or player and an \emph{advantage} feature.
Including this additional feature improves performance significantly, and we conclude that representing opponents in terms of match-dependent combinations of features can be very useful in practice.

\begin{table}[ht]
  \caption{
Predictive performance of models with a home or first-mover advantage in comparison to models without.}
  \label{tab:homeadv}
  \centering
  \sisetup{table-format=1.3, detect-all}
\begin{tabular}{l SS SS}
  \toprule
                  & \multicolumn{2}{c}{Basic}         & \multicolumn{2}{c}{Advantage} \\
                    \cmidrule(r){2-3}                   \cmidrule(r){4-5}
  Dataset         &            Loss &     \text{Acc.} &            Loss &     \text{Acc.}  \\
  \midrule
  World football  &           0.926 &           0.558 & \bfseries 0.900 & \bfseries 0.579  \\
  ChessBase small &           1.026 &           0.480 & \bfseries 1.019 & \bfseries 0.485  \\
  \bottomrule
\end{tabular}

\end{table}

\subsubsection{Capturing Intransitivity} \label{sec:eval-intrans}
\begin{figure}[t]
  \includegraphics{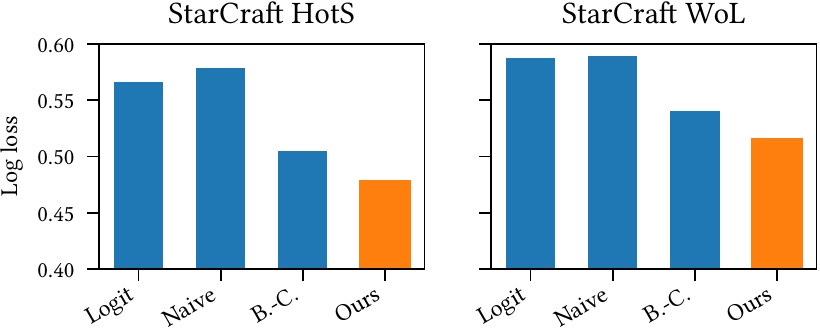}
  \caption{
Average log loss of four models (Bradley--Terry, naive, blade-chest and ours) on the StarCraft datasets.}
  \label{fig:starcraft}
\end{figure}

Score-based models such as ours are sometimes believed to be unable to capture meaningful intransitivities, such as those that arise in the ``rock-paper-scissors'' game~\citep{chen2016modeling}.
This is incorrect: if an opponent's score can be modeled by using match-dependent features, we can simply add an \emph{interaction} feature for every pair of opponents.
In the next experiment, we model the score difference between two opponents $i, j$ as $d \doteq s_i - s_j + s_{ij}$.
Informally, the model learns to explain the transitive effects through the usual player scores $s_i$ and $s_j$ and the remaining intransitive effects are captured by the interaction score $s_{ij}$.
We compare this model to the Blade-Chest model of \citet{chen2016modeling} on the two StarCraft datasets, known to contain strong intransitivities.
The Blade-Chest model is specifically designed to handle intransitivities in comparison data.
We also include two baselines, a simple Bradley--Terry model without the interaction features (\emph{logit}) and a non-parametric estimator (\emph{naive}) that estimates probabilities based on match outcomes between each pair---without attempting to capture transitive effects.
As shown in Figure~\ref{fig:starcraft}, our model outperforms all other approaches, including the Blade-Chest model.
More details on this experiment can be found in Appendix~\ref{app:intrans}.

\subsection{Inference Algorithm}
\label{sec:evalinf}
We turn our attention to the inference algorithm and study the impact of several implementation choices.
We start by quantifying the impact of the mean-field assumption~\eqref{eq:approxpost} and of the choice of variational objective on predictive performance.
Then, we demonstrate the scalability of the algorithm on the ChessBase dataset and measure the acceleration obtained by parallelizing the algorithm.

\subsubsection{Mean-Field Approximation}
In order to gain understanding on the impact of the factorization assumption in~\eqref{eq:approxpost}, we devise the following experiment.
We consider a small subset of the basketball data containing all matches between 2000 and 2005 ($N = 6382$, $M = 32$).
We evaluate the predictive performance on each week of the last season by using all the matches prior to the test week as training data.
Our model uses a one-to-one mapping between teams and features, a constant + Matérn 1/2 covariance function, and a Gaussian likelihood on the points difference.

We compare the predictive performance resulting from two inference variants,
\begin{enuminline}
\item mean-field approximate inference, i.e., Algorithm~\ref{alg:inference}, and
\item \emph{exact} posterior inference\footnote{%
This is possible for this particular choice of likelihood thanks to the self-conjugacy of the Gaussian distribution, but at a computational cost $O(N^3)$.}.
\end{enuminline}
Both approaches lead to an average log loss of \num{0.634} and an average accuracy of \num{0.664}.
Strikingly, both values are equal up to four decimal places, suggesting that the mean-field assumption is benign in practice \citep{birlutiu2007expectation}.

\subsubsection{Variational Objective}
Next, we study the influence of the variational method.
We re-run the experiments of Section~\ref{sec:evaldyn}, this time by using the reverse-KL objective instead of EP.
The predictive performance in terms of average log loss and average accuracy is equal to the EP case (Table~\ref{tab:predperf}, last three columns) up to three decimal places, for all four datasets.
Hence, the variational objective seems to have little practical impact on predictive performance.
As such, we recommend using the reverse-KL objective for likelihoods whose log-partition function~\eqref{eq:logpart} cannot be computed in closed form, as the numerical integration of the expected log-likelihood~\eqref{eq:exp-ll} is generally more stable.

\subsubsection{Scalability}
\begin{figure}[t]
  \includegraphics{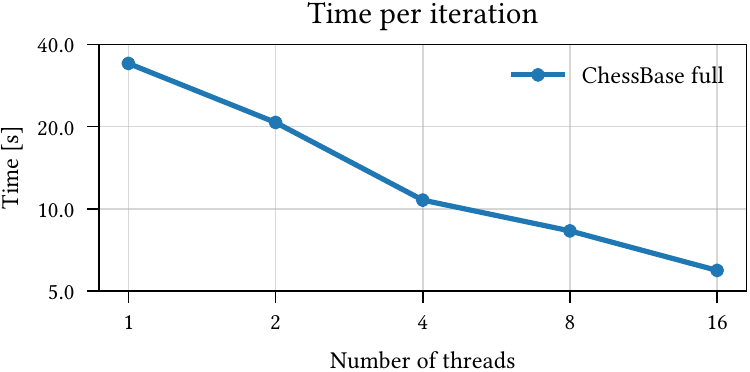}
  \caption{
Running time per iteration of a multithreaded implementation of Algorithm~\ref{alg:inference} on the ChessBase full dataset, containing over \num{7} million observations.}
  \label{fig:scalability}
\end{figure}

Finally, we demonstrate the scalability of our inference algorithm by training a model on the full ChessBase dataset, containing over \num{7} million observations.
We implement a multithreaded version of Algorithm~\ref{alg:inference} in the Go programming language\footnote{%
The code is available at \url{https://github.com/lucasmaystre/gokick}.}
and run the inference computation on a machine containing two 12-core Intel Xeon E5-2680 v3 (Haswell generation) processors clocked at 2.5 GHz.
Figure~\ref{fig:scalability} displays the running time per iteration as function of the number of worker threads.
By using \num{16} threads, we need only slightly over \num{5} seconds per iteration.

\section{Related Work}
\label{sec:relwork}

Probabilistic models for pairwise comparisons have been studied for almost a century.
\citet{thurstone1927law} proposed his seminal \emph{law of comparative judgment} in the context of psychology.
Almost concurrently, \citet{zermelo1928berechnung} developed a method to rank chess players from match outcomes.
Both rely on the same idea: objects are characterized by a latent score (e.g., the intrinsic quality of a perceptual variable, or a chess player's skill) and the outcomes of comparisons between objects depend on the difference between the corresponding latent scores.
Zermelo's model was later rediscovered by \citet{bradley1952rank} and is currently usually referred to as the Bradley--Terry model.
\citet{stern1992all} provides a unifying framework and shows that, in practice, Thurstone's and Zermelo's models result in similar fits to the data.
In the context of sports, some authors suggest going beyond ordinal outcomes and investigate pairwise-comparison models with Gaussian \citep{guo2012score}, Poisson \citep{maher1982modelling, guo2012score}, or Skellam \citep{karlis2009bayesian} likelihoods.

In many applications of practical interest, comparison outcomes tend to vary over time.
In chess, for example, this is due to the skill of players changing over time.
The World Chess Federation, which uses a variant of the Bradley--Terry model to rank players, updates player scores after each match by using a stochastic gradient update:
\begin{align*}
s_i \gets s_i + \lambda \frac{\partial}{\partial s_i} \log p(y \mid s_i - s_j),
\end{align*}
where $\lambda \in \mathbf{R}$ is a learning rate.
It is interesting that this simple online update scheme (known as the Elo rating system \citep{elo1978rating}) enables a basic form of ``tracking'': the sequence of scores gives an indication of a player's evolution over time.
Whereas, in this case, score dynamics occur as a by-product of the learning rule, several attempts have been made to model time dynamics explicitly.
Usually, these models assume a variant of Brownian motion:
\begin{align}
\label{eq:brownian}
s(t_{n+1}) = s(t_n) + \varepsilon_{n},
    \qquad \varepsilon_{n} \sim \mathcal{N}(0, \sigma^2 \Abs{t_{n+1} - t_n}).
\end{align}
\citet{glickman1993paired} and \citet{fahrmeir1994dynamic} are, to the best of our knowledge, the first to consider such a model.
\citet{glickman1999parameter} derives a computationally-efficient Bayesian inference method by using closed-form approximations of intractable integrals.
\citet{herbrich2006trueskill} and \citet{dangauthier2007trueskill} propose a similar method based on Gaussian filtering and expectation propagation, respectively.
\citet{coulom2008whole} proposes a method based on the Laplace approximation.
Our model strictly subsumes these approaches; Brownian motion is simply a special case of our model obtained by using the Wiener kernel.
One of the key contributions of our work is to show that it is not necessary to restrict the dynamics to Brownian motion in order to get linear-time inference.

Finally, we briefly review literature on the link between Gaussian processes (GPs) with scalar inputs and state-space models (SSMs), as this forms a crucial component of our fast inference procedure.
Excellent introductions to this link can be found in the theses of \citet{saatci2012scalable} and \citet{solin2016stochastic}.
The connection is known since the seminal paper of \citet{ohagan1978curve}, which introduced Gaussian processes as a method to tackle general regression problems.
It was recently revisited by \citet{hartikainen2010kalman}, who provide formulae for going back-and-forth between GP covariance and state-space forms.
Extensions of this link to non-Gaussian likelihood models are discussed in \citet{saatci2012scalable} and \citet{nickisch2018state}.
To the best of our knowledge, we are the first to describe how the link between GPs and SSMs can be used in the context of observation models that combine \emph{multiple} processes, by using a mean-field variational approximation.

\section{Conclusions}
\label{sec:conclusions}

We have presented a probabilistic model of pairwise comparison outcomes that can capture a wide range of temporal dynamics.
This model reaches state-of-the-art predictive performance on several sports datasets, and it enables generating visualizations that help in understanding comparison time-series.
To fit our model, we have derived a computationally efficient approximate Bayesian inference algorithm.
To the best of our knowledge, our algorithm is the first linear-time Bayesian inference algorithm for dynamic pairwise comparison models that minimizes the reverse-KL divergence.

One of the strengths of our approach is that it enables to discover the structure of the time dynamics by comparing the log-marginal likelihood of the data under various choices of covariance functions.
In the future, we would like to fully automatize this discovery process, in the spirit of the \emph{automatic statistician} \citep{duvenaud2014automatic}.
Ideally, given only the comparison data, we should be able to systematically discover the time dynamics that best explain the data, and generate an interpretable description of the corresponding covariance functions.

\begin{acks}
We thank Holly Cogliati-Bauereis, Ksenia Konyushkova, and the anonymous reviewers for careful proofreading and constructive feedback.
\end{acks}

\balance

\bibliographystyle{ACM-Reference-Format}
\bibliography{kickscore}

\clearpage
\appendix

\section{Inference Algorithm}
\label{app:inference}

For conciseness, we drop the index $n$ and consider a single observation $(\bm{x}, t^*, y) \in \mathcal{D}$.
Let $\bm{s} \doteq \bm{s}(t^*)$ be the vector containing the score of all features at the time of the observation.
Instead of optimizing the ``standard'' parameters $\tilde{\mu}_m, \tilde{\sigma}^2_m$, we will optimize the corresponding \emph{natural} parameters $\tilde{\alpha}_m, \tilde{\beta}_m$.
They are related through the following equations.
\begin{align*}
\tilde{\alpha}_m = \tilde{\mu}_m / \tilde{\sigma}^2_m, \qquad
\tilde{\beta}_m = 1 / \tilde{\sigma}^2_m
\end{align*}

\paragraph{Expectation Propagation}

Let $q_{-}(\bm{s}) = \mathcal{N}(\bm{\mu}, \bm{\Sigma})$ be the cavity distribution.
The log-partition function can be rewritten as a one-dimensional integral:
\begin{align*}
\log Z
    &\doteq \log \Exp[q_{-}]{ p(y \mid \bm{x}^\Tr \bm{s}) }
     = \log \int_{\bm{s}} p(y \mid \bm{x}^\Tr \bm{s}) \mathcal{N}(\bm{s} \mid \bm{\mu}, \bm{\Sigma}) d\bm{s} \\
    &= \log \int_{u} p(y \mid u) \mathcal{N}(u \mid \mu, \sigma^2) du,
\end{align*}
where $\mu = \bm{x}^\Tr \bm{\mu}$ and $\sigma^2 = \bm{x}^\Tr \bm{\Sigma} \bm{x}$.
The function \textsc{Derivatives} computes the first and second derivatives with respect to this mean, i.e.,
\begin{align*}
\delta_1 = \frac{\partial}{\partial \mu} \log Z, \qquad
\delta_2 = \frac{\partial^2}{\partial \mu^2} \log Z.
\end{align*}
Given these quantities, the function \textsc{UpdateParams} updates the pseudo-observations' parameters for each $m \in \mathcal{X}$:
\begin{align*}
\tilde{\alpha}_m &\gets (1 - \lambda) \tilde{\alpha}_m
    + \lambda \left[ \frac{x_m \delta_1 - \mu_m x_m^2 \delta_2}{1 + \Sigma_{mm} x_m^2 \delta_2} \right], \\
\tilde{\beta}_m  &\gets (1 - \lambda) \tilde{\beta}_m 
    + \lambda \left[ \frac{-x_m^2 \delta_2}{1 + \Sigma_{mm} x_m^2 \delta_2} \right],
\end{align*}
where $\lambda \in (0, 1]$ is a learning rate.
A formal derivation of these update equations can be found in \citet{seeger2007bayesian, rasmussen2006gaussian, minka2001family}.

\paragraph{Reverse KL Divergence}

Let $q(\bm{s}) = \mathcal{N}(\bm{\mu}, \bm{\Sigma})$ be the current posterior.
Similarly to the log-partition function, the expected log-likelihood can be rewritten as a one-dimensional integral:
\begin{align*}
L
    &\doteq \Exp[q]{ \log p(y \mid \bm{x}^\Tr \bm{s}) }
     = \int_{\bm{s}} \log p(y \mid \bm{x}^\Tr \bm{s}) \mathcal{N}(\bm{s} \mid \bm{\mu}, \bm{\Sigma}) d\bm{s} \\
    &= \int_{u} \log p(y \mid u) \mathcal{N}(u \mid \mu, \sigma^2) du,
\end{align*}
where $\mu = \bm{x}^\Tr \bm{\mu}$ and $\sigma^2 = \bm{x}^\Tr \bm{\Sigma} \bm{x}$.
The function \textsc{Derivatives} computes the first and second derivatives with respect to this mean, i.e.,
\begin{align*}
\delta_1 = \frac{\partial}{\partial \mu} L, \qquad
\delta_2 = \frac{\partial^2}{\partial \mu^2} L.
\end{align*}
Given these quantities, the function \textsc{UpdateParams} updates the pseudo-observations' parameters for each $m \in \mathcal{X}$:
\begin{align*}
\tilde{\alpha}_m &\gets (1 - \lambda) \tilde{\alpha}_m 
    + \lambda \left[ x_m \delta_1 - \mu_m x_m^2 \delta_2 \right], \\
\tilde{\beta}_m  &\gets (1 - \lambda) \tilde{\beta}_m
    + \lambda \left[ -x_m^2 \delta_2 \right],
\end{align*}
where $\lambda \in (0, 1]$ is the learning rate.
A formal derivation of these update equations can be found in \citet{khan2017conjugate}.

\section{Experimental Evaluation}
\label{app:eval}

The code used to produce the experiments presented in this paper is publicly available online.
It consists of two software libraries.
\begin{itemize}
\item A library written in the Python programming language, available at \url{https://github.com/lucasmaystre/kickscore}.
This libary provides a reference implementation of Algorithm~\ref{alg:inference} with a user-friendly API.

\item A library written in the Go programming language, available at \url{https://github.com/lucasmaystre/gokick}.
This library provides a multithreaded implementation of Algorithm~\ref{alg:inference}, focused on computational performance.
\end{itemize}
Additionally, the scripts and computational notebooks used to produce the experiments and figures presented in this paper are available at \url{https://github.com/lucasmaystre/kickscore-kdd19}.

\subsection{Hyperparameters}

Generally speaking, we choose hyperparameters based on a search over \num{1000} configurations sampled randomly in a range of sensible values (we always make sure that the best hyperparameters are not too close to the ranges' boundaries).
In the case of our models, we choose the configuration that maximizes the log-marginal likelihood of the training data.
In the case of TrueSkill and Elo, we choose the configuration that minimizes the leave-one-out log loss on the training data.

A list of all the hyperparameters is provided in Table~\ref{tab:hpdefs}.
A formal definition of the covariance functions we use is given in Table~\ref{tab:covfuncs}.
Finally, Table~\ref{tab:hpvals} lists the hyperparameter values used in most of the experiments described in Section~\ref{sec:eval}.

\begin{table}[ht]
  \caption{
Hyperparameters and their description.}
  \label{tab:hpdefs}
  \centering
  \renewcommand{\arraystretch}{1.2}
  \begin{tabular}{c l}
    \toprule
    Symbol                   & Description \\
    \midrule
    $\lambda$                & Learning rate                             \\
    $\alpha$                 & Draw margin                               \\
    $\sigma^2_n$             & Observation noise (Gaussian likelihood)   \\
    $\sigma_{\text{cst}}^2$  & Variance (constant covariance)            \\
    $\sigma_{\text{lin}}^2$  & Variance (linear covariance)              \\
    $\sigma_{\text{W}}^2$    & Variance (Wiener covariance)              \\
    $\nu$                    & Smoothness (Matérn covariance)            \\
    $\sigma_{\text{dyn}}^2$  & Variance (Matérn covariance)              \\
    $\ell$                   & Timescale, in years (Matérn covariance) \\
    \bottomrule
  \end{tabular}
\end{table}

\begin{table}[ht]
  \caption{
Covariance functions.}
  \label{tab:covfuncs}
  \centering
  \renewcommand{\arraystretch}{1.2}
  \begin{tabular}{l l}
    \toprule
    Name                & $k(t, t')$  \\
    \midrule
    Constant            & $\sigma_{\text{cst}}^2$  \\
    Linear              & $\sigma_{\text{lin}}^2 t t'$  \\
    Wiener              & $\sigma_{\text{W}}^2 \min \{ t, t' \}$  \\
    Matérn, $\nu = 1/2$ & $\sigma_{\text{dyn}}^2 \exp (-\Abs{t - t'} / \ell)$  \\
    Matérn, $\nu = 3/2$ & $\sigma_{\text{dyn}}^2 (1 +\sqrt{3} \Abs{t - t'} / \ell) \exp (-\sqrt{3} \Abs{t - t'} / \ell)$  \\
    \bottomrule
  \end{tabular}
\end{table}

\begin{table*}[t]
  \caption{
Hyperparameter values used for the models of Section~\ref{sec:eval}.}
  \label{tab:hpvals}
  \centering
  \sisetup{table-format=1.3}
\begin{tabular}{lll SSS[table-format=3.3]SSScS[table-format=2.3]S[table-format=3.3]}
  \toprule
  Dataset         &  Model  &  Likelihood  &  {$\lambda$}  &  {$\alpha$}  &  {$\sigma^2_n$}
                      &  {$\sigma_{\text{cst}}^2$} &  {$\sigma_{\text{lin}}^2$}  &  {$\sigma_{\text{W}}^2$}
                      &  {$\nu$}  &  {$\sigma_{\text{dyn}}^2$}  &  {$\ell$} \\
  \midrule        
  ATP tennis      & Constant                & Probit      & 1.000 &  \Emd &    \Emd & 0.817 &  \Emd &  \Emd & \Emd &   \Emd &   \Emd \\
                  & Elo                     & Logit       & 0.262 &  \Emd &    \Emd &  \Emd &  \Emd &  \Emd & \Emd &   \Emd &   \Emd \\
                  & TrueSkill               & Probit      &  \Emd &  \Emd &    \Emd & 0.137 &  \Emd & 0.007 & \Emd &   \Emd &   \Emd \\
                  & Ours                    & Probit      & 1.000 &  \Emd &    \Emd & 0.366 & 0.001 & 0.147 & \Emd &   \Emd &   \Emd \\
                  & Figure~\ref{fig:scores} & Probit      & 1.000 &  \Emd &    \Emd & 0.034 &  \Emd &  \Emd &  3/2 &  0.912 &  7.469 \\
  \midrule
  NBA Basketball  & Constant                & Probit      & 1.000 &  \Emd &    \Emd & 0.060 &  \Emd &  \Emd & \Emd &   \Emd &   \Emd \\
                  & Elo                     & Logit       & 0.095 &  \Emd &    \Emd &  \Emd &  \Emd &  \Emd & \Emd &   \Emd &   \Emd \\
                  & TrueSkill               & Probit      &  \Emd &  \Emd &    \Emd & 0.128 &  \Emd & 0.001 & \Emd &   \Emd &   \Emd \\
                  & Ours                    & Probit      & 1.000 &  \Emd &    \Emd & 0.003 &  \Emd &  \Emd &  1/2 &  0.152 &  3.324 \\
                  & Figure~\ref{fig:scores} & Probit      & 1.000 &  \Emd &    \Emd & 0.003 &  \Emd &  \Emd &  3/2 &  0.138 &  1.753 \\
                  & Table~\ref{tab:lklperf} & Logit       & 1.000 &  \Emd &    \Emd & 0.001 &  \Emd &  \Emd &  1/2 &  0.417 &  3.429 \\
                  & Table~\ref{tab:lklperf} & Gaussian    & 1.000 &  \Emd & 143.451 & 0.059 &  \Emd &  \Emd &  1/2 & 17.667 &  3.310 \\
                  & Table~\ref{tab:lklperf} & Poisson-exp & 0.800 &  \Emd &    \Emd & 5.470 &  \Emd &  \Emd &  1/2 &  0.003 &  2.378 \\
  \midrule
  World Football  & Constant                & Probit      & 1.000 & 0.372 &    \Emd & 0.933 &  \Emd &  \Emd & \Emd &   \Emd &   \Emd \\
                  & Elo                     & Logit       & 0.196 & 0.578 &    \Emd &  \Emd &  \Emd &  \Emd & \Emd &   \Emd &   \Emd \\
                  & TrueSkill               & Probit      &  \Emd & 0.381 &    \Emd & 1.420 &  \Emd & 0.001 & \Emd &   \Emd &   \Emd \\
                  & Ours                    & Probit      & 1.000 & 0.386 &    \Emd & 0.750 &  \Emd &  \Emd &  1/2 & 0.248 &  69.985 \\
                  & Table~\ref{tab:lklperf} & Logit       & 1.000 & 0.646 &    \Emd & 2.001 &  \Emd &  \Emd &  1/2 & 0.761 &  71.693 \\
                  & Table~\ref{tab:lklperf} & Gaussian    & 1.000 &  \Emd &   3.003 & 4.062 &  \Emd &  \Emd &  1/2 & 2.922 & 175.025 \\
                  & Table~\ref{tab:lklperf} & Poisson-exp & 0.800 &  \Emd &    \Emd & 0.300 &  \Emd &  \Emd &  1/2 & 0.210 &  83.610 \\
                  & Table~\ref{tab:homeadv} & Probit      & 1.000 & 0.407 &    \Emd & 0.895 &  \Emd &  \Emd &  1/2 & 0.220 &  44.472 \\
  \midrule
  ChessBase small & Constant                & Probit      & 1.000 & 0.554 &    \Emd & 0.364 &  \Emd &  \Emd & \Emd &  \Emd &    \Emd \\
                  & Elo                     & Logit       & 0.157 & 0.856 &    \Emd &  \Emd &  \Emd &  \Emd & \Emd &  \Emd &    \Emd \\
                  & TrueSkill               & Probit      &  \Emd & 0.555 &    \Emd & 0.240 &  \Emd & 0.001 & \Emd &  \Emd &    \Emd \\
                  & Ours                    & Probit      & 1.000 & 0.558 &    \Emd & 0.307 &  \Emd & 0.010 & \Emd &  \Emd &    \Emd \\
                  & Table~\ref{tab:homeadv} & Probit      & 1.000 & 0.568 &    \Emd & 0.188 &  \Emd &  \Emd &  1/2 & 0.188 &  35.132 \\
  \bottomrule
\end{tabular}

\end{table*}

\subsection{Capturing Intransitivity}
\label{app:intrans}

We closely follow the experimental procedure of \citet{chen2016modeling} for the experiment of Section~\ref{sec:eval-intrans}.
In particular, we randomly partition each dataset into three splits: a training set (\num{50}\% of the data), a validation set (\num{20}\%), and a test set (\num{30}\%).
We train the model on the training set, choose hyperparameters based on the log loss measured on the validation set, and finally report the average log loss on the test set.

For the Blade-Chest model, we were not able to reproduce the exact results presented in \citet{chen2016modeling} using the open-source implementation available at \url{https://github.com/csinpi/blade_chest}.
Instead, we just report the best values in Figures 3 and 4 of the paper (\emph{blade-chest inner} model, with $d = 50$).

For the Bradley--Terry model, we use the Python library \emph{choix} and its function \texttt{opt\_pairwise}.
The only hyperparameter to set is the regularization strength $\alpha$.

For our model, since there are no timestamps in the StarCraft datasets, we simply use constant covariance functions.
The two hyperparameters are $\sigma^2_{\text{cst}}$ and $\sigma^2_{\times}$, the variance of the player features and of the interaction features, respectively.
The hyperparameter values that we used are given in Table~\ref{tab:hpintrans}.

\begin{table}[ht]
  \caption{
Hyperparameter values for the experiment of Section~\ref{sec:eval-intrans}.}
  \label{tab:hpintrans}
  \centering
  \sisetup{table-format=1.3}
  \begin{tabular}{l S SS}
    \toprule
                    & \text{Bradley--Terry} & \multicolumn{2}{c}{Ours}  \\
                      \cmidrule(r){2-2}       \cmidrule{3-4}
    Dataset         & {$\alpha$}            & {$\sigma^2_{\text{cst}}$} & {$\sigma^2_{\times}$} \\
    \midrule
    StarCraft WoL   &                 0.077 &                     4.821 &                 3.734 \\
    StarCraft HotS  &                 0.129 &                     4.996 &                 4.342 \\
  \bottomrule
  \end{tabular}
\end{table}

\balance

\end{document}